\documentclass[twoside]{article}

\usepackage[accepted]{aistats2020}
\usepackage[round]{natbib}

\usepackage[utf8]{inputenc} % allow utf-8 input
\usepackage[T1]{fontenc}    % use 8-bit T1 fonts
\usepackage{hyperref}       % hyperlinks
\usepackage{url}            % simple URL typesetting
\usepackage{booktabs}       % professional-quality tables
\usepackage{amsfonts}       % blackboard math symbols
\usepackage{nicefrac}       % compact symbols for 1/2, etc.
\usepackage{microtype}      % microtypography

\usepackage{amssymb}
\usepackage{amsmath}
\usepackage{mathtools}
\usepackage{graphicx}
\usepackage{wrapfig}
\usepackage{subcaption}
\usepackage{hyperref}
\usepackage{tikz}
\usepackage{algorithm,algorithmic}
\newcommand{\NAME}{BMI}

\usepackage{tikz}
\newcommand*\circled[1]{\tikz[baseline=(char.base)]{
            \node[shape=circle,draw,inner sep=1pt] (char) {#1};}}

\newcommand\todo[1]{}
\newcommand\todop[1]{}

\DeclarePairedDelimiterX{\infdivx}[2]{(}{)}{%
  #1\;\delimsize\|\;#2%
}

\newcommand{\KL}{\text{KL}\infdivx}
\newcommand{\p}{\mathbb{P}}
\newcommand{\q}{Q}
\newcommand{\qw}{\omega}
\newcommand{\tk}{t}

\begin{document}

\runningtitle{Better Long-Range Dependency
  By Bootstrapping A Mutual Information Regularizer}


\twocolumn[

\aistatstitle{Better Long-Range Dependency \\
  By Bootstrapping A Mutual Information Regularizer}
% \aistatsauthor{}
 \aistatsauthor{Yanshuai Cao$^*$ \And Peng Xu$^*$ }
% \aistatsaddress{} ]
\aistatsaddress{ Borealis AI } ]

\begin{abstract}
     In this work, we develop a novel regularizer to improve the learning of long-range dependency of sequence data. Applied on language modelling, our regularizer expresses the inductive bias that sequence variables should have high mutual information even though the model might not see abundant observations for complex long-range dependency. We show how the ``next sentence prediction (classification)" heuristic can be derived in a principled way from our mutual information estimation framework, and be further extended to maximize the mutual information of sequence variables. The proposed approach not only is effective at increasing the mutual information of segments under the learned model but more importantly, leads to a higher likelihood on holdout data, and improved generation quality. Code is released at \url{https://github.com/BorealisAI/BMI}.
\end{abstract}

%!TEX root = ./acl2019.tex

\section{Introduction}

Transformer-based large scale pre-training \citep{devlin2018bert,yang2019xlnet,zhang2019ernie,sun2019ernie,radford2019language} has yielded impressive successes in many NLP tasks. Among the many components introduced by BERT \citep{devlin2018bert} originally, the auxiliary task of next sentence prediction (NSP) is regarded as a heuristic, which is actually a binary classification task to distinguish if another sentence is the correct next sentence or a randomly sampled sentence from the corpus. As an ad-hoc heuristic, NSP is often dropped by some subsequent works \citep{joshi2019spanbert,liu2019roberta} on large scale pre-training based on empirical performance, but is picked up in other NLP problems \citep{xu2019cross,liu2019comes}. This work explores a hidden connection of NSP to mutual information maximization, providing a more principled justification for those applications where NSP is used. The new insight is independent of the transformer architecture, and it allows us to design a new algorithm that shows additional improvements beyond NSP for RNN language modelling, in terms of improving long-range dependency learning.

Learning long-range dependency in sequential data such as text is
challenging, and the difficulty has mostly been attributed to the
vanishing gradient problem in autoregressive neural networks such as
RNNs \citep{hochreiter2001gradient}. There is a vast literature trying
to solve this gradient flow problem through better architecture
\citep{hochreiter2001gradient,mikolov2014learning,vaswani2017attention},
better optimization \citep{martens2011learning} or better
initialization \citep{le2015simple}. On the other hand, there is an
orthogonal issue that has received less attention: statistical
dependency over a short span is usually abundant in data, e.g.,
bigrams, common phrases and idioms; on the other hand, long-range
dependency typically involves more complex or abstract relationships
of a large number of tokens (\emph {high order interactions}). In
other words, there is a sampling mismatch between observations
supporting local correlations versus evidence for high order
interaction, while the latter requires more samples to learn from at
the first place because they involve more variables. 
%(although compositionality could alleviate the increased sample complexity to some extent). 
We conjecture that in addition to the gradient flow issue, this
problem of sparse sampling of high order statistical relations renders
learning long-range dependency hard in natural language processing.

Take language modelling for example: with a vocabulary of size $K$,
the number of possible sequences grows as $K^m$ with sequence length
$m$. Neural language models use distributed representation to overcome
this issue \citep{bengio2003neural}, as not all $K^m$
sequences form plausible natural language utterances, and there is
shared semantics and compositionality in different texts. 
However, the parametrization does not change the fundamental fact that
in the training data, there is an abundance of observation for local
patterns, but much sparser observations for the different high-level
ideas.  As language evolved to express the endless possibilities of the world,
even among the set of ``plausible'' long sequences, a training set can
only cover a small fraction.
%Indeed, a sufficiently large corpus could potentially cover almost all plausible bigrams on fixed finite vocabulary, but never all possible thoughts.
Therefore, there is an inherent imbalance of sampling between short and long range dependencies. As such,
because it is a data sparsity issue at the core, it cannot be
completely solved by better architecture or optimization.

The natural remedy facing limited data is to regularize the model
using prior knowledge. 
In this work, we propose a novel approach for
incorporating into the usual maximum likelihood objective the
additional prior that long-range dependency exists in texts.
We achieve this by bootstrapping a lower bound on the mutual information
(MI) over groups of variables (segments or sentences) and subsequently
applying the bound to encourage high MI. The first step of bootstrapping the lower bound
is exactly the NSP task.
Both the bootstrapping and
application of the bound improves long-range dependency learning:
first, the bootstrap step helps the neural network's hidden
representation to recognize evidence for high mutual information that
exists in the \emph{data distribution}; second, the information lower
bound value as the reward encourages the \emph{model distribution} to
exhibit high mutual information as well.
We apply the proposed method for language modelling, although the
general framework could apply to other problems as well.

%% The later would require policy gradient reinforcement learning, but we
%% find a sample-efficient alternative via a novel modified version of
%% Reward Augmented Maximum Likelihood (RAML) \cite{norouzi2016reward}.

Our work offers a new perspective on why the heuristic of
next sentence prediction used in previous works
\citep{trinh2018learning,devlin2018bert} are useful auxiliary
tasks, while revealing missing ingredients, which we complete in the
proposed algorithm. We demonstrate improved perplexity on two established benchmarks, reflecting the positive regularizing effect.
We also show that our proposed method can help the model generate
higher-quality samples with more diversity measured by reversed
perplexity \citep{zhao2018adversarially} and more dependency measured by an empirical lower bound of mutual information.

\section{Background}
%\vspace{-.2cm}
\subsection{MLE Language Model and Sparsely Observed High Order Dependency}
%\vspace{-.1cm}
A language model (LM) assigns a probability to a sequence of tokens
(characters, bytes, or words). Let $\tau_i$ denote token
variables, a LM $\q$ factorizes the joint distribution of
$\tau_i$'s into a product of conditionals from left to right, leveraging
the inherent order of text $\q(\tau_1, \ldots, \tau_k) = \prod\nolimits_{i=1}^{k} \q(\tau_i | \tau_{< i})$,
where $\tau_{< i}$ denotes all token variables with index less than $i$, and $\q(\tau_1 | \tau_{< 1}) = \q(\tau_1)$.
Let $(\tk_i)^n_{i=1}$ be an observed sequence of tokens as training data, sampled from \emph{data distribution} $\p$. Learning simply maximizes the log likelihood of the observations with respect to the parameters $\qw$ of $\q$ (we will use the notation $\q$ and $\q_\qw$ interchangeably.):
%%\vspace{-.1cm}
\begin{equation}
\mathit{L}_{\text{MLE}}(\qw) = \sum\nolimits_{i=1}^{n} \log \q_\qw(\tau_i = \tk_i | \tk_{< i}) \label{mle}
\end{equation}

As $\mathit{L}_{\text{MLE}}$ requires $\q$ to focus its probability
mass on {\it observed} subsequent tokens given its preceding ones,
maximum likelihood does have the ability to enforce long-range
dependencies of sequence variables. However, Eq.\ \ref{mle} hides
issues about high order interactions where a relatively smaller fraction of the valid outcomes are observed.
To see this, take a partition of the sequence variables $(\tau_i)^n_{i=1}$ into $[\tau_{< a}, X, Y]$, where $X = (\tau_{a}, \ldots, \tau_{b})$, and $Y = (\tau_{b+1}, \ldots, \tau_{n})$, then Eq.\ \ref{mle} is equivalent to:
\begin{align*}
   \mathit{L}_{\text{MLE}}(\qw) &= \sum\nolimits_{i=1}^{b} \log \q_\qw(\tau_i = \tk_i | \tk_{< i}) \\
   &+ \log \q_\qw(Y\!\!=\!\!(\tk_{b+1}, \ldots, \tk_{n})| X\!\!=\!\!(\tk_{a}, \ldots, \tk_{b}), \tk_{< a})
\end{align*}
Now we can see that as in the case of a single next token prediction, MLE
prefers $\q$ to commit its prediction to the particular observed
sequence(s) of $Y$, but this observed set is too sparse for the much
larger configuration space.
We propose to use MI as a way to express the belief that there is some dependency between $X$ and $Y$ without committing to particular instantiated predictions.

\subsection{Regularizing Mutual Information}
%\vspace{-.1cm}
Mutual information (MI) is a measure of how much does observing one random variable reveal about another (and vice versa). It is zero if and only if the two are independent.
The MI $I(X;Y)$ between two random variables $X$ and $Y$ (scalars or vectors) is the Kullback-Leibler (KL) divergence between the joint $\p_{XY}$ and product of marginal distributions $\p_X \otimes \p_Y$ of the two random variables:
%\vspace{-.1cm}
 \begin{equation}
  I(X; Y) = \KL{\p_{XY}}{\p_X \otimes \p_Y} \label{mi_kl_def}
 \end{equation}%
 For text data, $X$ and $Y$ can be sentences or segments of tokens (potentially extending over sentence boundaries).
 As MI is defined with respect to the distribution,
 rather than the particular observed values, it enables us to enforce
 dependency without committing to instantiated predictions.
 
We can also write $I(X; Y)$ as the difference between entropy and
conditional entropy:
%\vspace{-.1cm}
\begin{equation}
  I(X; Y) = H(Y) - H(Y|X) = H(X) - H(X|Y) \label{MI_X}
\end{equation}%
Hence, high MI can be achieved by minimizing conditional entropy or maximizing marginal entropy (or both). Unlike MLE which can only maximize MI by reducing the conditional entropy, a MI regularizer has the option to encourage long-range dependency without forcing $\q$ to commit its prediction to observed sequence(s), but by increasing the marginal entropy $H(Y)$. 

Note that the definition in Eq.\ \ref{mi_kl_def} and Eq.\ \ref{MI_X}
depend on the distribution used, so under the data and model
distributions ($\p$ and $\q$), the MI is not the same in
general. Henceforth, we will make the distinction of $I^\p$ and $I^\q$
in our notations. 

$I^\p$ cannot be directly computed due to lack of functional form of $\p$. For RNN or Transformer based autoregressive models, evaluating $I^\q$ is computationally intractable since it needs summation over all possible sequences. Hence, we will instead lower bound $I^\p$ and $I^\q$ in a computationally tractable way.

%!TEX root = ./neurips_2019.tex
\section{Boostrapping a Mutual Information Regularizer}
%\vspace{-.1cm}
\begin{figure*}[h]  
  \begin{subfigure}[l]{.96\columnwidth}
      \centering
  \includegraphics[width=\columnwidth]{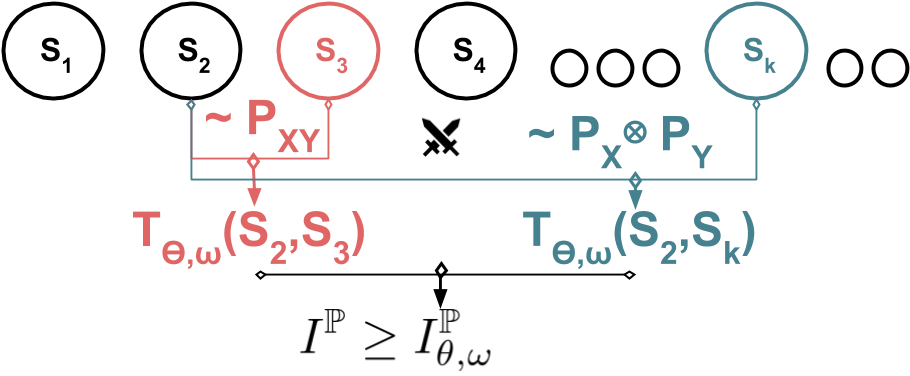}
  \caption{Mutual information lower bound: 
    learn to classify the correct next sentence
    from a randomly sampled one: essentially the next
    sentence prediction task, which was previously considered a heuristic \citep{devlin2018bert}.} \label{fig_I_p}
  \end{subfigure}%
  \hfill 
  \begin{subfigure}[r]{.96\columnwidth}
      \centering
      \includegraphics[width=\columnwidth]{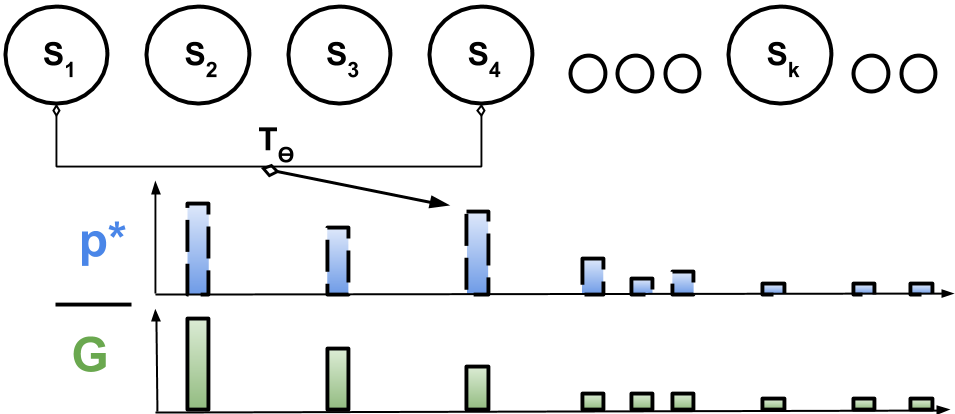}
      \caption{Importance-Weighted RAML: sample another nearby
        sentence ($S_4$), and maximize the
        conditional log likelihood of it given $S_1$ but with an
        appropraite weight, which is calculated using the MI estimator
        from Fig.\ \ref{fig_I_p}. }\label{fig_raml}
  \end{subfigure}
  \caption{Overview of the two key components of the proposed approach}
  %\vspace{-.cm}
\end{figure*}

\todop{Better to be summarized into one paragraph}

Our operating assumption is that longer segments in
the data should have high $I^\p$ with each other; and our goal is for sequence
variables under model $Q$ to have similarly high $I^\q$.

At a high level, our method adds some regularization terms to the
MLE objective Eq.\ \ref{mle}, in two separate phases. The illustration
in Fig.\ \ref{fig_I_p}-\ref{fig_raml} capture the core of our
proposal. 
In the first
phase, we bootstrap a MI lower bound by doing
next sentence prediction, which is a binary classification of the correct next sentence versus a randomly sampled sentence. After some switching condition is met, we proceed to the second phase where the MI estimator is also used to produce reward for optimizing $I^\q$ directly using reward augmented maximum likelihood.

In order to compute the proposed regularizers, we add a small discriminator net (parametrized by $\theta$) on top of the base model $\q$'s hidden features (parametrized by $\qw$). The discriminator will then look at pairs of segments or sequence, the $S$'s in Fig.\ \ref{fig_I_p}, trying to distinguish pairs following some joint distribution ($S$'s with dependency) versus product of marginals (independent $S$'s).

The discriminator serves the MI regularization in both
phases. For the first phase, Sec.\ \ref{sec_I_p_bound} will show that making this bound
tight automatically forces the hidden representation of $\q$ to
preserve as much MI as possible, making the model $\q$ good at
recognizing related information.
After $\q$ and discriminator are sufficiently well trained, the learned parameters $(\theta,\qw)$ can then be applied to MI under $\q$ distribution, to get a lower bound $I^\q_{\theta,\qw} \leq I^\q$. This leads to the second phase, where in addition to continue to optimize $I^\p_{\theta,\qw}$, we use $I^\q_{\theta,\qw}$ as reward to encourage high MI under $\q$. This has a more direct regularizing effect than $I^\p_{\theta, \qw}$.

Directly optimizing $I^\q_{\theta,\qw}$ requires sampling from $\q$ and learning by policy gradient (or other gradient estimators). However, sequential sampling from $Q$ is slow while deep RL converges slowly due to high variance. Hence, we explore an alternative, the reward augmented maximum likelihood (RAML) \citep{norouzi2016reward}. Because RAML does not directly support our MI bound as the reward, we develop a modification via importance reweighting in Sec.\ref{sec_iw_raml}. The overall algorithm is summarized in Alg.\ \ref{algo}.

\subsection{Phase-I: Next Sentence Prediction Bootstraps a Lower Bound of $I^\p(X;Y)$}
\label{sec_I_p_bound}
%\vspace{-.1cm}
As previously mentioned, $I^\p$ cannot be directly computed, but can be lower bounded in a number of ways, for example, via the MINE lower bound \citep{belghazi2018mine} $I^\p(X; Y) \geq I^\p_\zeta(X, Y)$:
%\vspace{-.2cm}
\begin{equation}
  I^\p_\zeta(X, Y) = E_{\p_{XY}}(T_\zeta(X,Y)) -\log E_{\p_X{\otimes}\p_Y}(e^{T_\zeta(X,Y)}) \label{MINE}  
\end{equation}
where $T_\zeta(X,Y)$ is a parametrized test function trying to
distinguish samples of the joint distribution from those from the product
of marginals. %\todo{what is the range of $T_\zeta$?}
$T_\zeta(X,Y)$ can be any function and optimizing $\zeta$ makes the bound tighter. Hence, we compose some intermediary hidden layer representation $\phi_\qw(.)$ of the neural net (e.g. RNN or transformer) with a discriminator $D_\theta: \Phi \rightarrow \mathbb{R}$, in order to form the test function $T_\zeta(X,Y)=T_{\theta, \qw}(X,Y)$:
%\vspace{-.1cm}
\begin{equation}
T_{\theta, \qw}(X,Y)=D_\theta(\phi_\qw(X), \phi_\qw(Y))
\end{equation}
For brevity, we will write $\phi^X_\qw = \phi_\qw(X)$ and $\phi^Y_\qw= \phi_\qw(Y)$ henceforth.
% For the regularizer, the encodings are performed separately for the two sentences/segments, without passing the RNN hidden states from $X$ to $Y$ as in the language model part.
% The reason for this requirement will become clear in the next section.

In this work, we take $X$ and $Y$ of $\p_{XY}$ to be consecutive pair of
sentences. Other pairs could also be regularized in theory, such
as consecutive segments, or pairs of sentences at special positions in
a document, like the first sentence of consecutive paragraphs.
\todop{Last sentence is unnecessary?}

Eq.\ \ref{MINE} can be optimized using noise contrastive estimation, by turning it into a
binary classification problem as in \cite{hjelm2018learning}. To sample positive examples from
$\p_{XY}$, we draw $X = S_l$ for some sentence indexed $l$ and $Y =
S_{l+1}$, $(X, Y) = (S_l, S_{l+1})$. To sample negatives from the product
of marginals $\p_X\otimes\p_Y$, we take $X = S_l$, and sample $Y = S_k$ where $S_k$ randomly drawn from the training corpus. Fig.\ \ref{fig_I_p}
depicts our overall approach to bootstrap this lower bound. 
As pointed out by \cite{hjelm2018learning}, when the goal is to maximize the MI rather than estimating its particular value, one can use a proxy $\tilde{I}^\p_{\theta,\qw}$ that has
better gradient property than $I^\p_{\theta,\qw}$: \todo{are negatives sampled uniformly from the corpus or within a context window?}
\begin{align}
  \tilde{I}^\p_{\theta,\qw} = &E_{\p_{XY}}[-\text{SP}(-D_\theta(\phi^X_\qw,\phi^Y_\qw))] \nonumber\\
&- E_{\p_X{\otimes}\p_Y}[\text{SP}({D_\theta(\phi^X_\qw,\phi^Y_\qw)})]
  \label{p_mi_term}
\end{align}
where $\text{SP}(x) = \log(1\!+\!e^x)$. $I^\p_{\theta,\qw}$ remains a lower bound for any parameters.

%%  in this work, we can see that the ``next sentence
%% prediction'' where a classifier distinguishes the correct next
%% sentence from a  task can be viewed as estimating a MI lower bound.

\subsubsection{Regularizing Effect on Model $\q$}
\label{Ip_effect_on_q}
%\vspace{-.1cm}
To understand how does maximizing $I^\p_{\theta, \qw}$ regularize the model $\q$, note that the MI between the encodings is a lower bound on the MI of the raw inputs, by the Data Processing Inequality (DPI) \citep{cover2012elements}. In other words, $I^\p(X;Y) \geq I^\p(\phi^X_\qw; \phi^Y_\qw)$, which can be proved in a straightforward way by applying the DPI twice: $I^\p(X;Y) \geq I^\p(X; \phi^Y_\qw) \geq I^\p(\phi^X_\qw;\phi^Y_\qw)$.
The first inequality hold due to the DPI applied on the markov chain $X \rightarrow Y \rightarrow \phi(Y)$; then the second one on $\phi(Y) \rightarrow X \rightarrow \phi(X)$. 
Note that the Markov chains are not additional assumption, but merely a statement that
$\phi(X)$ does not dependent on $Y$ when $X$ is given (similarly for
the first Markov chain).

Because $D_\theta$ is also the test function for the joint versus product of marginals on the random variables $\phi^X_\qw$ and $\phi^Y_\qw$, we have $I^\p(X;Y) \geq I^\p(\phi^X_\qw; \phi^Y_\qw) \geq I^\p_\theta (\phi^X_\qw, \phi^Y_\qw) = I^\p_{\theta, \qw} (X, Y)$, i.e. \emph{ the MI of features is sandwiched between the MI of data and our parametric lower bound $I^\p_{\theta, \qw}$.} 

Therefore, while $I^\p(X;Y)$ is a fixed value for the data, estimating a bound for $I^\p$ by optimizing both $\theta$ and $\qw$ pushes the hidden representation to capture as much data MI as possible.
Viewed from a different angle, it is equivalent to estimating a bound for the MI between $\phi^X_\qw$ and $\phi^Y_\qw$,
$I^\p(\phi^X_\qw; \phi^Y_\qw)$ (using the add-on discriminator $D_\theta$), and then optimize the $Q$-model features
$\phi^X_\qw$ and $\phi^Y_\qw$ to have high mutual
information. 

Intuitively, this step encourages $\phi_\qw$'s to be good representations of inputs that
recognize related information in the data. However, the MI of data $I^\p(X;Y)$ is a property of the data (distribution) $\p$, not of the model $\q$ afterall. If the encoder is already very powerful, i.e. $I^\p(\phi^X_\qw; \phi^Y_\qw)$ already close to $I^\p(X;Y)$, the sandwiching effect from the lower bound would not be significant. This is consistent with observations of the recent works \citep{joshi2019spanbert,liu2019roberta,yang2019xlnet} which drop NSP based on lack of empirical improvements.
However, the theoretical connection to MI implies that we need to maximize $I^\q$, which NSP (Phase-I) is not directly doing. In the next section, we will develop a method to directly optimize $I^\q$.

\subsection{Phase-II: Directly Optimizing $I^\q(X,Y)$}
%\vspace{-.1cm}
As mentioned, the regularization effect of Phase-I is indirect, as the expectation is with respect to the data distribution $\p$. We now discuss how to directly and efficiently optimize $I^\q(X,Y)$.

%and apply it to yield a lower bound $I^\q \geq I^\q_{\theta,\qw}$

To this end, after sufficient training from Phase-I, we take the learned parameters $\theta,\qw$ to initialize the lower bound $I^\q_{\theta,\qw}$. Optimizing $I^\q_{\theta,\qw}$ poses a series of challenges which we will tackle in the next subsections (Sec.\ \ref{sec_q_rl}-\ref{sec_iw_raml}). We emphasize that during Phase-II, we still optimize $I^\p_{\theta,\qw}$ from Phase-I, but just with an additional regularization term, which together approximate for $I^\q_{\theta,\qw}$.

\subsubsection{Difficulty with optimizing $I^\q_{\theta,\qw}$}
\label{sec_q_rl}
%\vspace{-.1cm}
%% \todo{How to do this without referring to the proxy objective? Maybe
%%   just say due to product rule of differentiating, the first term has
%%   gradient with respect to MI reward; the second one has is policy gradient. }

Because the MINE bound holds for any parameters, we can instead use the binary classification form to optimize the parameters, similar to what we do for $I^\p_{\theta,\qw}$ and as done in \cite{hjelm2018learning}.
The proxy objective has the form: $\tilde{I}^\q_{\theta,\qw} = E_{\q_{XY}} R^+_{\theta,\qw} - E_{{\q}_X{\otimes}{\q}_Y} R^-_{\theta,\qw}$ where,
\begin{align}
  R^+_{\theta,\qw}  &= -\text{SP}(-D_\theta(\phi^X_\qw,\phi^Y_\qw)) \\
R^-_{\theta,\qw}  &=  \text{SP}({D_\theta(\phi^X_\qw,\phi^Y_\qw)})  
\end{align}
To optimize $\tilde{I}^\q_{\theta,\qw}$ with respect to $\zeta = (\theta, \qw)$, the gradient has two terms $\nabla_\zeta \tilde{I}^\q_{\theta,\qw} = g_1 + g_2$, where
\begin{align}
 g_1 &= E_{\q_{XY}} \nabla R^+_{\theta,\qw} - E_{{\q}_X{\otimes}{\q}_Y} \nabla R^-_{\theta,\qw} \label{g1}\\
 g_2 &= E_{\q_{XY}} R^+_{\theta,\qw} \nabla \log \q_{XY} \nonumber \\
 &\phantom{=}- E_{{\q}_X{\otimes}{\q}_Y} R^-_{\theta,\qw}  (\nabla \log {\q}_X+\nabla \log {\q}_Y) \label{g2}
\end{align}
$g_2$ uses policy gradient (i.e. likelihood ratio estimator) with $\q$ being the policy while $R^+$ and $R^-$ being the reward (and penalty). $g_2$ can be variance-reduced by control-variate methods, e.g.\ \cite{rennie2017self}.

However, deep RL is known to converge slowly due to high variance, our
trials confirm the difficulty in this particular case. Furthermore,
sampling from $\q$ is generally slow for autoregressive models as it
cannot be easily parallelized. These two issues compounded means that
we would like to avoid sampling from $\q$.
To this end, we develop a modification of the reward augmented maximum likelihood (RAML) \citep{norouzi2016reward}, which avoids the high variance and slow $\q$-sampling.
\todop{Reviewers may be confused about how to sample from $Q$ with the conditional or the marginal}

For the $g_1$ part (Eq.\ \ref{g1}), if we simply replace the $\q$ distributions with $\p$ in the expectation, we recover the Phase-I regularizer Eq.\ \ref{p_mi_term}, which we can use to approximate $g_1$. The bias of this approximation is:
 %\vspace{-.2cm}
\begin{align}
  &\sum\nolimits_{X,Y} (\q(X,Y) - \p(X,Y)) \nabla R^+ \nonumber \\
   &-\sum\nolimits_{X,Y} (\q(X)\q(Y) - \p(X)\p(Y)) \nabla R^-
\end{align}
which becomes small as the maximum likelihood learning progresses, because in both terms, the total variation distance $\sum |\q - \p|$ is bounded by $\sqrt{2 \KL{\p}{\q}}$ via Pinsker's inequality \citep{Tsybakov:2008:INE:1522486}.

\subsubsection{IW-RAML: RAML background}
%\vspace{-.1cm}
RAML can be viewed as optimizing the reverse direction of KL divergence comparing to the entropy-regularized policy gradient RL objective. We will leave the details of RAML to the Appendix.\ \ref{app_raml_background} and refer readers to the work \citep{norouzi2016reward}. For our purpose here, the important information is that the RAML gradient with the policy gradient are:
\begin{align}
\nabla L_{\text{\tiny{RAML}}} &=- E_{p^\star_\beta(Y|Y^\star)}
\left\{\nabla \log \q_\qw(Y | X)\right\} \label{raml_grad} \\
\nabla L_{\text{\tiny{RL}}} &=- E_{\q_\qw(Y | X)}\left\{ r(Y, Y^\star) \nabla \log \q_\qw(Y | X)\right\}\label{rl_grad}
\end{align}
where $p^\star_\beta(Y|Y^\star)$ is the \emph{exponentiated pay-off distribution} defined as:
\begin{equation}
p^\star_\beta(Y | Y^\star) = {\exp\{r(Y, Y^\star) / \beta\}} \big/ {Z(Y^\star, \beta)} \label{payoff_distribution}
\end{equation}
$r(Y, Y^\star)$ is a reward function that measures some similarity of
$Y$ with respect to the ground truth $Y^\star$ (e.g. negative
edit-distance). RAML gradient Eq.\ \ref{raml_grad} samples from a stationary distribution, while policy gradient Eq.\ \ref{rl_grad} samples from the changing $\q_\qw$ distribution. Furthermore, by definition, samples from $p^\star_\beta(Y|Y^\star)$ has higher chance for high reward, while samples $\q_\qw(Y | X)$ relies on exploration. For these reasons, RAML has much lower variance than RL.

\subsubsection{IW-RAML: MI Reward}
\label{sec_iw_raml}
%\vspace{-.1cm}
Unfortunately, sampling from $p^\star_\beta(Y|Y^\star)$ can only be done efficiently for some special classes of reward such as the edit-distance used in \cite{norouzi2016reward}. Here, we would like to use the learned MI estimator, more specifically the classifier scores as the reward. Assume $Y^\star$ is the sentence following $X$ in the corpus, then for any other $Y$, the reward is:
\begin{equation}
r(Y, Y^\star;X)\!=\!D_\theta(\phi^X_\qw,\phi^Y_\qw)\!-\!D_\theta(\phi^X_\qw,\phi_\qw(Y^\star)) \label{raml_reward}
\end{equation}
In the illustration Fig.\ \ref{fig_raml}, $X$ would be $S_1$ and $Y^\star=S_2$, and another $Y = S_4$ is sampled to be evaluated. $Y$ could also be any other sentence/segment not in the dataset.

As the deep-neural-net-computed scores lack the simple
structure of edit-distance that can be exploited for efficient
sampling from $p^\star_\beta(Y|Y^\star)$, direct application of RAML to the MI reward is not possible. We will instead develop an efficient alternative based on importance sampling.

\todop{The specific choice of $G$ can be put into Appendix}
Intuitively, a sentence that is near $X$ in the text would tend to be more related to it, and vice versa. Therefore, we can use a geometric distribution based at the index of $Y^\star$ as the proposal distribution, as illustrated in Fig.\ \ref{fig_raml}. Let $Y^\star$ have sentence/segment index $m$, then
\begin{equation}
G(Y = S_k | Y^\star=S_m) = (1-\lambda) ^ {(k - m)} \lambda 
\end{equation}
where $\lambda$ is a hyperparameter (we set to $.3$ without tuning it). Other proposals are also possible.
With $G$ as the proposal, our \emph{importance weighted RAML} (IW-RAML) gradient is then:
\begin{equation}
\nabla L_{\text{RAML}} = - E_{G} \left( \nabla \log \q_\qw(Y | X) {p^\star_\beta(Y|Y^\star)} \big/ {G(Y|Y^\star)}\right)\label{iw_raml_grad}
\end{equation}
Because the reward in Eq.\ \ref{raml_reward} is shift-standardized with respect to the discriminator score at $Y^\star$, we assume that the normalization constant $Z$ in Eq.\ \ref{payoff_distribution} does not vary heavily for different $Y^\star$, so that we can perform self-normalizing importance sampling by averaging across the mini-batches. 

\subsubsection{IW-RAML: Bias-Variance Trade-off}
%\vspace{-.1cm}
\todop{Maybe summarized in one or two sentences. Put the detailed discussions into Appendix.}

A side benefit of introducing $G$ is to re-establish the stationarity of the sampling distribution in the RAML gradient estimator. Because the reward function Eq.\ \ref{raml_reward} depends on $(\theta,\qw)$, the exponentiated pay-off distribution is no longer stationary like in the original RAML with simple reward \citep{norouzi2016reward}, but we re-gain stationarity through the fixed proposal $G$, keeping the variance low. Stationarity of the sampling distribution is one of the reasons for the lower variance in RAML.

Choosing IW-RAML over RL is a bias-variance trade-off. The RL objective gradient in Eq.\ \ref{g1}-\ref{g2} is the unbiased one, and IW-RAML as introduced has a few biases: using the opposite direction of the KL divergence (analyzed in \cite{norouzi2016reward}); distribution support of $G$ being smaller than $p^\star_\beta(Y|Y^\star)$. Each of these approximations introduces some bias, but the overall variance is significantly reduced as the empirical analysis in Sec.\ \ref{sec_exp_var} shows.

%% Note that because the reward $r(Y, Y^\star; X)$ is over all possible
%% sentence/segments, not constrained to the training dataset, our
%% proposal actually does not cover the full support of $p^\star_\beta(Y|Y^\star)$, which introduces a
%% bias. However, restricted to the sentences/segments in the corpus, $G$ is a
%% reasonably good proposal, and the sampling space is much smaller, so
%% the variance is very low.

%\todop{didn't get a clear idea about what does ``support" mean; overall the part about support is a bit confusing, better to elaborate or drop}
%Given enough computation resources and tuning, using the deep RL approach of Sec.\ \ref{sec_q_rl} within our bootstrapped MI regularization framework could potentially yield good result too.

\begin{algorithm}
  \scriptsize
  %\footnotesize
  \caption{Language Model Learning with \NAME\ regularizer}
  \label{algo}
\begin{algorithmic}[1]
\STATE {\bf Input:} batch size $M$, dataset $\Omega$, proposal distribution $G$, maximum number of iterations $N$.
\STATE{phase-two := \FALSE}
\FOR{$\text{itr}=1,\dots,N$}
\STATE Compute LM objective $\mathit{L}_{\text{MLE}}(\qw)$ from Eq.~\ref{mle} and its gradient; \# \circled{1}\\
\STATE Sample a mini-batch of consecutive sentences $\{X_i, Y_i\}_1^M$ from $\Omega$ as samples from $\p_{XY}$;
\STATE Sample another mini-batch of $\{Y_i^{-}\}_1^M$ from $\Omega$ to form $\{X_i, Y^{-}_i\}_1^M$ as samples from $\p_{X}\otimes\p_{Y}$;\\
\STATE Extract features $\phi_\qw^{X}$, $\phi_\qw^{Y}$ and $\phi_\qw^{Y^{-}}$ and compute $\tilde{I}^\p_{\theta,\qw}$ according to Eq.~\ref{p_mi_term} and its gradient; \# \circled{2}
\IF{phase-two}
\STATE Sample a mini-batch of $\{\tilde Y_i\}_1^M$ from $\Omega$ according to $G$, each with corresponding $Y^\star = Y_i$.
\STATE Compute IW-RAML gradients according to Eq.~\ref{iw_raml_grad}, with $Y^\star = Y_i$, $Y = \tilde Y_i$, and $X = X_i$. \# \circled{3}
\ENDIF
\STATE Add gradient contributions from \circled{1}, \circled{2}, \circled{3} and update parameters $\qw$ and $\theta$.
\IF{\NOT phase-two \AND meeting switch condition}
\STATE phase-two := \TRUE
\ENDIF
\ENDFOR
\end{algorithmic}
\end{algorithm}

\section{Related Work}
\vspace{-.2cm}
\textbf{Long Range Dependency and Gradient Flow~~} Capturing long-range dependency has been a major challenge in sequence learning. Most works have focused on the gradient flow in backpropagation through time (BPTT). The LSTM architecture \citep{lstm1997} was invented to address the very problem of vanishing and exploding gradient in RNN \citep{hochreiter2001gradient}. There is a vast literature on improving the gradient flow with new architectural modification or regularization \citep{mikolov2014learning,koutnik2014clockwork,wu2016multiplicative,li2018independently}. Seq-to-seq with attention or memory \citep{bahdanau2014neural,cho2015describing,sukhbaatar2015end,joulin2015inferring} is a major neural architecture advance that improves the gradient flow by shortening the path that relevant information needs to traverse in the computation graph. The recent invention of the Transformer architecture \citep{vaswani2017attention}, and the subsequent large scale pre-training successes \citep{devlin2018bert,radford2018improving,gpt2} are further examples of better architecture improving gradient flow.

\textbf{Regularization via Auxiliary Tasks~~} Closer to our method
are works that use auxiliary prediction tasks as regularization
\citep{trinh2018learning,devlin2018bert}. \cite{trinh2018learning} uses
an auxiliary task of predicting some random future or past subsequence
with reconstruction loss. Their focus is still on vanishing/exploding
gradient and issues caused by BPTT. Their method is justified
empirically and it is unclear if the auxiliary task losses are
compatible with maximum likelihood objective of language modelling,
which they did not experiment on. \cite{devlin2018bert} adds a ``next
sentence prediction'' task to its masked language model objective,
which tries to classify if a sentence is the correct next one or
randomly sampled. This task is the same as our Phase-I for learning
the lower bound $I^\p_{\theta, \qw}$, but we are the first to draw the
theoretical connection to mutual information, explaining its
regularization effect on the model (Sec.\ \ref{Ip_effect_on_q}), and
applying the bootstrapped MI bound for more direct regularization in
Phase-II is completely novel in our method.
%% Finally, the ``next
%% sentence prediction'' in BERT \cite{devlin2018bert} is done from the
%% feature corresponding to a special token ``[CLS]'', rather than from
%% all inputs' features, which has a weaker MI regularization effect
%% according to our theory in Sec.\ \ref{Ip_effect_on_q}. \todo{This
%%   might not be exactly accurate; the feature of CLS also depends on
%%   every token in the sentence.}

\textbf{Language Modeling with Extra Context~~} 
Modeling long range dependency is crucial to language models, since capturing the larger context effectively can help predict the next token.
In order to capture this dependency, there are some works that feed an additional representation of larger context into the network including additional block, document or corpus level topic or discourse information \citep{mikolov2012context, wang2015larger,dieng2016topicrnn, wang2017topic}. Our work is orthogonal to them and can be combined. 

%% \paragraph{MI Lower Bound and Deep Infomax}
%% \todo{Talk about MINE and feature learning via deep infomax}

%!TEX root = ./neurips_2019.tex
%\vspace{-.2cm}
\section{Experiments}
\label{sec_exp}
%\vspace{-.2cm}

\begin{table*}[ht]
  %\begin{minipage}{\textwidth}
    \centering
   %\small
{\caption{Perplexity and reverse perplexity on PTB and WT2.}}
\label{lm_ppl}
\begin{tabular}{l l l c l l c l l c l l}
\hline
& \multicolumn{5}{c}{PTB} && \multicolumn{5}{c}{WT2} \\ \cline{2-6} \cline{8-12}
& \multicolumn{2}{c}{PPL} && \multicolumn{2}{c}{Reverse PPL} && \multicolumn{2}{c}{PPL} && \multicolumn{2}{c}{Reverse PPL} \\ \cline{2-3} \cline{5-6} \cline{8-9} \cline{11-12}
Model & Valid & Test && Valid & Test && Valid & Test && Valid & Test \\ \hline
{\bf\scriptsize AWD-LSTM-MoS} & 58.08 & 55.97 && 82.88 & 77.57 && 66.01 & 63.33 && 93.52 & 88.79 \\ 
{\bf \NAME-base} & 57.16 & 55.02 && 80.64 & 75.31 && 64.24 & 61.67 && 90.95 & 86.31  \\ 
{\bf \NAME-full} & {\bf 56.85} & {\bf 54.65} && {\bf 78.46}  & {\bf 73.73} && {\bf 63.86} & {\bf 61.37}&& {\bf 90.20} & {\bf 85.11} \\
\hline \hline
{\bf\scriptsize AWD-LSTM-MoS (ft.)} & 56.54 & 54.44 && 80.29 & 75.51 && 63.88 & 61.45 && 91.32 & 85.69 \\ 
{\bf \NAME-base (ft.)} & 56.05 & 53.97 && 78.04 & 73.35 && 63.14 & 60.61 && 89.09 & 84.01 \\ 
{\bf \NAME-full (ft.)} & {\bf 55.61} & {\bf 53.67} && {\bf 75.81} & {\bf 71.81} && {\bf 62.99} & {\bf 60.51} && {\bf 88.27} & {\bf 83.43}\\
\hline
\end{tabular}
  %\end{minipage}
  %\vspace{-.3cm}  
\end{table*}

We experiment on two widely-used benchmarks on word-level language modeling, Penn Treebank (PTB) \citep{mikolov2012context} and WikiText-2 (WT2) \citep{merity2016pointer}.
We choose the recent state-of-the-art model among RNN-based models on these two benchmarks, AWD-LSTM-MoS \citep{yang2017breaking} as our baseline.
\todo{These results are not SOTA anymore?}

We compare the baseline with the same model adding variants of our proposed regularizer, Bootstrapping Mutual Information (\NAME) regularizer:
(1) {\bf \NAME-base}: apply Phase-I throughout the training;
(2) {\bf \NAME-full}: apply Phase-I till we learn a good enough $D_\theta$ then apply both Phase-I and Phase-II. Here, we adopt the same switching condition from SGD to ASGD \citep{polyak1992acceleration} in training RNN language model firstly proposed by \cite{merity2017regularizing} to switch from Phase-I to Phase-II.

%% \subsection{Experimental Setup}
%% \label{sec_setup}
%% %\vspace{-.2cm}

\begin{figure*}[ht]
  \centering
  \begin{subfigure}[b]{.8\columnwidth}
  \includegraphics[width=\columnwidth]{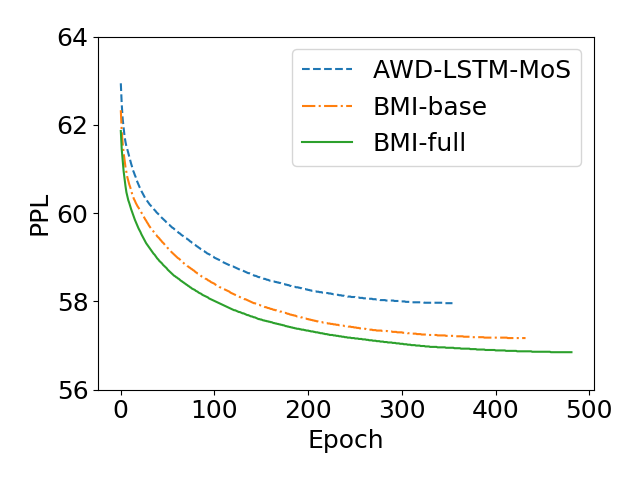}
  \caption{PTB}
  \end{subfigure}
    \centering
  \begin{subfigure}[b]{.8\columnwidth}
      \includegraphics[width=\columnwidth]{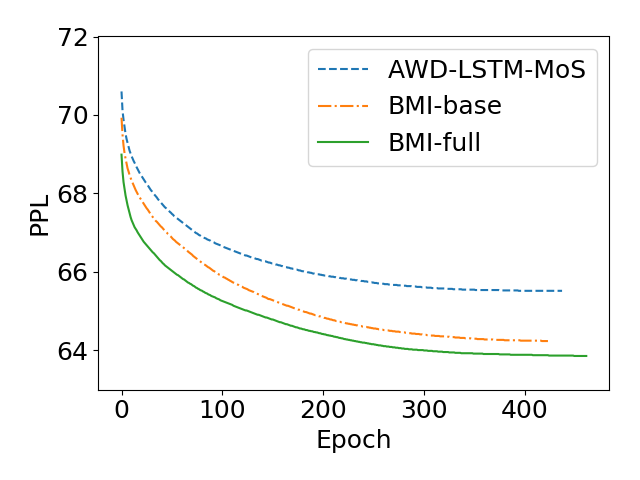}
      \caption{WT2}
  \end{subfigure}
  \caption{Learning curve for validation perplexity on PTB and WT2 after switching.} \label{curve}
  %\vspace{-.3cm}
\end{figure*}

\begin{table}[]
%\small
\centering
\captionof{table}{Estimated MI (lower bounds) of $X$ and $Y$, two random segments of length $40$ separated by $10$ tokens. Estimations using $10$-fold cross-validation and testing.}
\label{lm_ppl}
\begin{tabular}{l l l}
\hline
Generations & PTB & WT2 \\ \hline
{\scriptsize AWD-LSTM-MoS} & 0.25 $\!\pm\!$ 0.03 & 0.76 $\!\pm\!$ 0.03\\
{ \NAME-base} & 0.47 $\!\pm\!$ 0.03 & 0.88 $\!\pm\!$ 0.05\\
{ \NAME-full} & 0.48 $\!\pm\!$ 0.03 & 1.01 $\!\pm\!$ 0.06\\
{ Real Data} & 1.18 $\!\pm\!$ 0.08 & 2.14 $\!\pm\!$ 0.07\\
\hline
\end{tabular}
%\vspace{-.1cm}
\end{table}

\begin{figure}[]
\centering
 \includegraphics[width=.7\columnwidth]{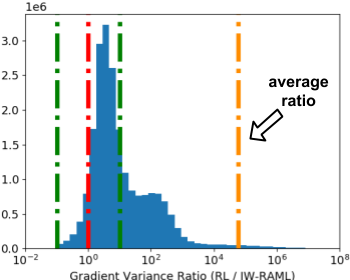}
 \captionof{figure}{Grad variance ratio (RL $/$ IW-RAML). Red dotted line indicates the ratio of 1, greens indicate the ratio of 0.1 and 10, orange indicates the average ratio of RL against IW-RAML.} \label{fig_ratio}
\end{figure}

\textbf{Experimental Setup~} We apply the max-pooling over the hidden states for all the layers in LSTM and concatenate them as our $\phi_\qw$-encoding.
We use a one-layer feedforward network with the features similar to \cite{conneau-EtAl:2017:EMNLP2017} as $[\phi^X_\qw, \phi^Y_\qw, \phi^X_\qw - \phi^Y_\qw, |\phi^X_\qw - \phi^Y_\qw|, \phi^X_\qw * \phi^Y_\qw]$ for our test function $D_\theta$ whose number of hidden units is $500$.
The ADAM \citep{kingma2014adam} optimizer with learning rate $2e^{-4}$ and weight decay of $1e^{-6}$ is applied on $\theta$, while $\qw$ is optimized in the same way as in \cite{merity2017regularizing,yang2017breaking} with SGD then ASGD \citep{polyak1992acceleration}.
All the above hyperparameters are chosen by validation perplexity on PTB and applied directly to WT2.
The weight of the regularizer term is set to $0.1$ for PTB and $0.02$ for WT2 chosen by validation perplexity on their respective datasets.
The remaining architecture and hyperparameters follow exactly the same as the code released by \cite{yang2017breaking}. As mentioned previously, we set the temperature hyperparameter $\beta$ in RAML to $1$, and $\lambda$ hyperparameter of importance sample proposal $G$ to $.3$, both without tuning. \todo{@peng Are the hyperparameters still these?}

All experiments are conducted on single (1080Ti) GPUs with PyTorch. 
We manually tune the following hyperparameters based on validation perplexity: the BMI regularizer weights in $[0.01, 0.02, 0.05, 0.1, 1.]$; $D_\theta$ hidden state size from $[100, 300, 500, 1000]$, Adam learning rate from $[1e-3, 2e-4]$.

%\vspace{-.2cm}
\subsection{Perplexity and Reverse Perplexity}
%\vspace{-.2cm}

Table \ref{lm_ppl} presents the main results of language modeling.
We evaluate the baseline and variants of our approach with and without finetune described in the baseline paper \citep{yang2017breaking}.
In all settings, the models with \NAME\ outperforms the baseline, and \NAME-full (with IW-RAML) yields further improvement on top of \NAME-base (without IW-RAML).

Following \cite{zhao2018adversarially}, we use {\em reverse perplexity} to measure the diversity aspect of generation quality.
We generate a chunk of text with $6M$ tokens from each model, train a second RNN language model (RNN-LM) on the generated text; then evaluate the perplexity of the held-out data from PTB and WikiText2 under the second language model. Note that the second RNN-LM is a regular LM trained from scratch and used for evaluation only. 
As shown in Table \ref{lm_ppl}, the models with \NAME\ regularizer improve the reverse perplexity over the baseline by a significant margin, indicating better generation diversity, which is to be expected as \emph{MI regularizer encourages higher marginal entropy} (in addition to lower conditional entropy).

Fig.~\ref{curve} shows the learning curves of each model on both datasets after switching to ASGD as mentioned earlier in Experiment Setup.
The validation perplexities of \NAME\ models decrease faster than the baseline AWD-LSTM-MoS.
In addition, {\NAME-full} is also consistently better than {\NAME-base} and can further decrease the perplexity after {\NAME-base} and AWD-LSTM-MoS stop decreasing.

%\begin{figure}[ht]
%\begin{center}
% \includegraphics[height=2.4in]{figures/penn_curve}
%\end{center}
%\caption{Learning curve for validation perplexity on PTB after switching.}
%\end{figure}
%
%\begin{figure}[ht]
%\begin{center}
% \includegraphics[height=2.4in]{figures/wiki_curve}
%\end{center}
%\caption{Learning curve for validation perplexity on WT2 after switching.}
%\end{figure}

\subsection{Empirical MI on generations}
%\vspace{-.2cm}
To verify that \NAME\ indeed increased $I^\q$, we measure the sample MI of generated texts as well as the training corpus. MI of long sequence pairs cannot be directly computed from samples, we instead estimate lower bounds by learning evaluation discriminators, $D_{\text{eval}}$ on the generated text. $D_{\text{eval}}$ is completely separate from the learned model, and is much smaller in size. We train $D_{\text{eval}}$'s using the proxy objective in Eq.\ \ref{p_mi_term} and early-stop based on the MINE lower bound Eq.\ \ref{MINE} on validation set, then report the MINE bound value on the test set. This estimated lower bound essentially measures the degree of dependency.
Table \ref{lm_ppl} shows that \NAME\ generations exhibit higher MI than those of the baseline AWD-LSTM-MoS, while \NAME-full improves over \NAME-base.

%\begin{wraptable}{t}{.55\textwidth}

%\end{wraptable}%

\subsection{Analysis: RL vs. IW-RAML variance}
\label{sec_exp_var}
%\vspace{-.2cm}
%\begin{wrapfigure}{r}{0.55\textwidth}

%\end{wrapfigure}%

Fig.~\ref{fig_ratio} compares the gradient variance under RL and IW-RAML on PTB. The gradient variance for each parameter is estimated over $200$ iterations after the initial learning stops and switches to ASGD; the ratio of variance of the corresponding parameters is then aggregated into the histogram. For RL, we use policy gradient with self-critical baseline for variance reduction \citep{rennie2017self}.
Only gradient contributions from the regularizers are measured, while the language model MLE objective is excluded. 

The histogram shows that the RL variance is more than $10^4$ times
larger than IW-RAML on average, and almost all of the parameters having higher gradient variance under RL. A
significant portion also has $1$-$4$ orders of magnitude higher
variance under RL than under IW-RAML. For this reason, policy gradient
RL does not contribute to learning when applied in Phase-II in our trials. \todo{exactly what number?}

%Our proposed IW-RAML reduce the variance significantly over policy gradient which is infeasible to use in practice due to high variance, while preserving the same global extremum.

%\begin{table}[h]
%\centering
%\small
%\begin{tabular}{l l l c l l}
%\hline
%& \multicolumn{2}{c}{PPL} && \multicolumn{2}{c}{Reverse PPL} \\ \cline{2-3} \cline{5-6}
%Model & Valid & Test && Valid & Test \\ \hline
%{\scriptsize AWD-LSTM-MoS} & 58.08 & 55.97 && 80.90 & 76.29 \\
%\NAME-base & 56.95 & 54.86 && 78.74 & 74.32 \\
%\NAME-full & {\bf 56.62} & {\bf 54.58} && {\bf 78.68} & {\bf 74.09} \\
%\hline
%{\scriptsize AWD-LSTM-MoS (ft.)} & 56.54 & 54.44 && 80.23 & 75.15 \\
%\NAME-base (ft.) & 55.75 & 53.72 && 76.05 & 72.06 \\
%\NAME-full (ft.) & {\bf 55.51} & {\bf 53.37} && {\bf 75.33} & {\bf 71.13} \\
%\hline
%\end{tabular}
%\caption{Perplexity and reverse perplexity on validation and test sets on PTB. ``(ft.)" indicates the finetune step in \citet{merity2017regularizing,yang2017breaking}.}
%\label{ptb_ppl}
%\end{table}
%\vspace{-.6cm}
%\begin{table}[h]
%\centering
%\small
%\begin{tabular}{l l l c l l}
%\hline
%& \multicolumn{2}{c}{PPL} && \multicolumn{2}{c}{Reverse PPL} \\ \cline{2-3} \cline{5-6}
%Model & Valid & Test && Valid & Test \\ \hline
%{\scriptsize AWD-LSTM-MoS} & 66.01 & 63.33 && 94.60 & 89.41 \\
%\NAME-base & 65.00 & 62.40 && 91.61 & 85.96 \\
%\NAME-full & {\bf 64.59} & {\bf 62.01} && {\bf 89.83} & {\bf 84.59} \\ 
%\hline
%\end{tabular}
%\caption{Perplexity and reverse perplexity on validation and test sets on WT2.}
%\label{wiki_ppl}
%\end{table}

\section{Conclusion}
\vspace{-.2cm}
We have proposed a principled mutual information regularizer for improving
long-range dependency in sequence modelling. The work also provides more principled explanation for the next sentence prediction (NSP) heuristic, but improves on it with a method for directly maximizing the mutual information of sequence variables. Finally, driven by this new connection, a number of possible extensions for future works are possible. For example, encouraging high MI between the title, the first sentence of a paragraph, or the first sentence of an article, with the other sentences in the same context.

%% To the best of our knowledge, this is the first work to recognize and address the sparse sampling of high order interactions as an issue hindering long-range dependency learning, orthogonal from the gradient flow problem. 

\subsubsection*{Acknowledgements}

We thank all the anonymous reviewers for their valuable inputs.

\bibliographystyle{acl_natbib}
\bibliography{ref}
\clearpage
\newpage
\appendix
\section{Appendix}

%% \subsection{$I^\p(X;Y) \geq I^\p(\phi^X_\qw; \phi^Y_\qw)$}
%% \label{proof_dpe}
%% Proof: We apply the Data Processing Inequality (DPI) \cite{cover2012elements} twice:
%% $I^\p(X;Y) \geq I^\p(X; \phi^Y_\qw) \geq I^\p(\phi^X_\qw;\phi^Y_\qw)$.
%% The first inequality hold due to the DPI applied on the markov chain $X \rightarrow Y \rightarrow \phi(Y)$; then the second one on $\phi(Y) \rightarrow X \rightarrow \phi(X)$.

%% Note: the Markov chains are not additional assumption, but merely a statement that
%% $\phi(X)$ does not dependent on $Y$ when $X$ is given (similarly for
%% the first Markov chain).

\subsection{RAML Background}
\label{app_raml_background}

The key idea behind RAML is to observe that the entropy-regularized
policy gradient RL objective $L_{\text{RL}}$ can be written as (up to constant and
scaling):
%  L_{\text{\tiny{MLE}}}&=\sum\nolimits_{(X, Y^\star) \in \mathit{D}}\KL{\delta(Y|Y^\star)}{\q_\qw(Y | X)} \label{mle_loss}\\
\begin{equation}
  L_{\text{\tiny{RL}}}=\sum\nolimits_{(X, Y^\star) \in \mathit{D}} \KL{\q_\qw(Y | X)}{p^\star_\beta(Y|Y^\star)} \label{rl_loss}
\end{equation}
%where $\delta(Y|Y^\star) = 1$ if and only if $Y = Y^\star$ (delta
%distribution); and $p^\star_\beta(Y|Y^\star)$ is the
%\emph{exponentiated pay-off distribution} defined as:
where $p^\star_\beta(Y|Y^\star)$ is the \emph{exponentiated pay-off distribution} defined as:
\begin{equation}
p^\star_\beta(Y | Y^\star) = {\exp\{r(Y, Y^\star) / \beta\}} \big/ {Z(Y^\star, \beta)} \label{payoff_distribution}
\end{equation}
$r(Y, Y^\star)$ is a reward function that measures some similarity of
$Y$ with respect to the ground truth $Y^\star$ (e.g. negative
edit-distance).
%% The difference between Eq.\ \ref{mle_loss} and Eq.\ \ref{rl_loss} is
%% two fold: forward vs.\ reverse KL divergence; and delta
%% vs.\ exponentiated payoff distribution.
Whereas in RAML \cite{norouzi2016reward}, one optimizes the KL in the
reverse direction:
\begin{equation}
L_{\text{RAML}} =\sum\nolimits_{(X, Y^\star) \in \mathit{D}} \KL{p^\star_\beta(Y|Y^\star)}{\q_\qw(Y | X)} \label{raml_loss}
\end{equation}
It was shown that these two losses have the same global extremum and when away from it their gap is bounded under some conditions \cite{norouzi2016reward}.
Compare the RAML gradient with the policy gradient:
\begin{align}
\nabla L_{\text{\tiny{RAML}}} &=- E_{p^\star_\beta(Y|Y^\star)}
\left\{\nabla \log \q_\qw(Y | X)\right\} \label{raml_grad} \\
\nabla L_{\text{\tiny{RL}}} &=- E_{\q_\qw(Y | X)}\left\{ r(Y, Y^\star) \nabla \log \q_\qw(Y | X)\right\}\label{rl_grad}
\end{align}
RAML gradient samples from a stationary distribution, while policy gradient samples from the changing $\q_\qw$ distribution. Furthermore, samples from $p^\star_\beta(Y|Y^\star)$ has higher chance of landing in configurations of high reward by definition, while samples $\q_\qw(Y | X)$ relies on random exploration to discover sequences with high reward. For these reasons, RAML has much lower variance than RL.

%\subsection{Gradient of $I^\q$}
%% \label{grad_iq}

%% \begin{align}
%%  \nabla I^\q_\zeta(X, Y) &= \nabla E_{\q_{XY}}(T_\zeta(X,Y)) - \nabla
%%  \log E_{\q_X{\otimes}\q_Y}(e^{T_\zeta(X,Y)}) \\
%%  &=  E_{\q_{XY}}(\nabla T_\zeta(X,Y)) + E_{\q_{XY}}\left(T_\zeta(X,Y)
%%  \nabla \log \q_{XY} \right) \\
%%  &\phantom{=} -
%%  \frac{E_{\q_X{\otimes}\q_Y}\nabla{e^{T_\zeta(X,Y)}}}{E_{\q_X{\otimes}\q_Y}(e^{T_\zeta(X,Y)})}
%%  - \frac{E_{\q_X{\otimes}\q_Y}{e^{T_\zeta(X,Y)}}(\nabla\log\q_X + \nabla\log\q_Y)}{E_{\q_X{\otimes}\q_Y}(e^{T_\zeta(X,Y)})}\\
%% \end{align}

%% \subsection{Additional Experiment Details}
%% \label{add_exp_details}
%% All experiments are conducted on single (1080Ti) GPUs with PyTorch. 

%% We manually tune the following hyperparameters based on validation perplexity: the BMI regularizer weights in $[0.01, 0.02, 0.05, 0.1, 1.]$; $D_\theta$ hidden state size is chosen from $[100, 300, 500, 1000]$, Adam learning rate from $[1e-3, 2e-4]$.

%% \todo{Describe compute infrastructure, i.e. single 1080TI...}

\end{document}